\documentclass[10pt, onecolumn]{article}

\usepackage[utf8]{inputenc}
\usepackage{amsmath}
\usepackage{array}
\usepackage{graphicx}
\usepackage[hyphens]{url}
\usepackage{multirow}
\usepackage{hyperref}
\usepackage{float}
\usepackage{rotating}
\usepackage{bm}
\usepackage{xcolor}
\usepackage{subfig}
\usepackage{multicol}
\usepackage{authblk}

\usepackage{algorithm} 
\floatname{algorithm}{Algorithm}

\begin{document}
%
\title{Automatic Handgun Detection Alarm in Videos Using Deep Learning}
%
%
%


\author[1]{Roberto Olmos}
\author[1]{Siham Tabik}
\author[1,2]{Francisco~Herrera}
\affil[1]{Soft Computing and Intelligent Information Systems
research group} \affil[2]{Department of Computer Science and
Artificial Intelligence, University of Granada, 18071 Granada,
Spain. emails: siham@ugr.es, herrera@decsai.ugr.es}

\providecommand{\keywords}[1]{\textbf{\textit{Index terms---}} #1}

\maketitle

\begin{abstract}
Current surveillance and control systems still require human
supervision and intervention. This work presents a novel automatic
handgun detection system in videos appropriate for both,
surveillance and control purposes. We reformulate this detection
problem into the problem of minimizing false positives and solve it
by building the key training data-set guided by the results of a
deep Convolutional Neural Networks (CNN) classifier, then assessing
the best classification model under two approaches, the sliding
window approach and region proposal approach. The most promising
results are obtained by Faster R-CNN based model trained on our new
database. The best detector show a high potential even in low
quality youtube videos and provides  satisfactory results as
automatic alarm system. Among  30 scenes, it successfully activates
the alarm after five successive true positives in less than 0.2
seconds, in 27 scenes. We  also define a new metric,  Alarm
Activation per Interval (AApI), to assess the performance of a
detection model as an automatic detection system in videos.

\end{abstract}

\keywords{Classification, Detection, Deep learning, Convolutional
Neural Networks (CNNs), Faster R-CNN, VGG-16, Alarm Activation per
Interval}

%

%
%
%
%
\section{Introduction}
The crime rates caused by guns are very concerning in many places in
the world, especially in countries where the possession of guns is
legal or was legal for a period of time. The last statistics
reported by the United Nations Office on Drugs and Crime (UNODC)
reveals that the number of crimes involving guns per 100,000
habitants are very high in many countries, e.g., 21.5 in Mexico, 4.7
in United States and 1.6 in Belgium~\cite{CITA1}. In addition,
several psychological studies demonstrated that the simple fact of
having access to a gun increases drastically the probability of
committing a violent behavior.

One way to reducing this kind of violence is prevention via early
detection so that the security agents or policemen can act. In
particular, one innovative solution to this problem is to equip
surveillance or control cameras with an accurate automatic handgun
detection alert system. Related studies address the detection of
guns but only on X-ray or millimetric wave images and only using
traditional machine learning
methods~\cite{PREMIO3,PREMIO4,PREMIO5,PREMIO6,PREMIO7}.

In the last five years, deep learning in general and Convolutional
Neural Networks (CNNs) in particular have achieved superior results
to all the classical  machine learning methods in image
classification, detection and segmentation in several
applications~\cite{Le2013,Hinton2012,Sainath2013,ghazi2017plant,yu2016exploiting,shu2016computational}.
Instead of manually selecting features, deep learning CNNs
automatically discover increasingly higher level features from
data~\cite{krizhevsky2012imagenet,guo2016deep}. We aim at developing
a good gun detector in videos using CNNs.

A proper training of deep CNNs, which contain millions of
parameters, requires very large datasets, in the order of millions
of samples, as well as High Performance Computing (HPC) resources,
e.g., multi-processor systems accelerated with GPUs. Transfer
learning through fine-tuning is becoming a widely accepted
alternative to overcome these constraints. It consists of
re-utilizing the knowledge learnt from one problem to another
related one~\cite{pan2010survey}. Applying transfer learning with
deep CNNs depends on the similarities between the original and new
problem and also on the size of the new training set.

In general, fine-tuning the entire network, i.e., updating all  the
weights, is only used when the new dataset is large enough, else the
model could suffer overfitting especially among the first layers of
the network. Since these layers extract low-level features, e.g.,
edges and color, they do not change significantly and can be
utilized for several visual recognition tasks. The last layers of
the CNN  are gradually adjusted to the particularities of the
problem and extract high level features, which are not readable by
the human eye. 
In this work we used a VGG-16 based classification model pre-trained
on the ImageNet dataset (around 1.28 million images over 1,000
generic object classes)~\cite{simonyan2014very} and fine-tuned on
our own dataset of 3000 images of guns taken in  a variety of
contexts.

Using CNNs to automatically detect pistols in videos faces several
challenges:
\begin{itemize}
\item Pistols can be handled  with one or two hands in different ways and thus a large part
of the pistol can be occluded.
\item The process of designing a new
dataset is manual and time consuming.
\item The labeled dataset can not be re-utilized by different detection approaches since they require
different preprocessing and labeling operations and cannot learn
from the same labeled databases.
\item Automatic pistol detection alarm requires
the activation of the alarm in real time and only when the system is
 confident about the existence of a pistol in the scene.
\item Automatic detection alarm systems  require an accurate location of the pistol in the monitored scene.
\end{itemize}

As far as we know, this work presents the first automatic gun
detection alarm system that uses deep CNNs based detection models.
We focus on the most used type of handguns in
crimes~\cite{zawitz1995guns}, pistol, which includes, revolver,
automatic and semi-automatic pistols, six-gun shooters, horse pistol
and derringers. To guide the design of the new dataset and to find
the best detector we consider the following steps:
\begin{itemize}
\item we reformulate the problem of automatic pistol detection alarm
in videos into the problem of minimizing the number of false
positives where {\it pistol} represents the true class and
\item we evaluate and compare the VGG-16 based classifier using two
different detection approaches, the sliding window and region
proposals approaches.
\end{itemize}

Due to the particularities  of each approach, we applied different
optimizations  in each case. We evaluated increasing the number of
classes in the sliding window approach and designing a richer
training dataset for the region proposals approach.

As this work focuses on near real time solutions, we selected the
most accurate and fastest detector and assess its performance on
seven  videos of different characteristics. Then, we evaluated its
suitability as automatic pistol detection alarm system using a new
metric, the Alarm Activation Time per Interval (AATpI), that
measures the activation time for each scene with guns.


 The main contributions of this work are:
\begin{itemize}
\item Designing a new labeled database that makes the learning model achieve high detection qualities. Our experience in building the new dataset and detector can be useful to guide developing the solution of other different problems.

\item Finding the most appropriate CNN-based detector that achieves real-time pistol detection in videos.
\item Introducing a new metric, AATpI, to
 assess the suitability of the proposed detector as automatic detection alarm system.
\end{itemize}

From the experiments we found that the most promising results are
obtained by Faster R-CNN based model trained on our new database.
The best performing model shows a high potential even in low quality
youtube videos and provides satisfactory results as automatic alarm
system. Among 30 scenes, it successfully activates the alarm, after
five successive true positives, within an interval of time smaller
than 0.2 seconds, in 27 scenes.

 This paper is organized as follows. Section~2
gives a brief analysis of the most related papers. Section~3
provides an overview of the CNN model used in this work. Section~4
describes the procedure we have used to find the best detector that
reaches good precisions and low false positives rate. Section~5
analyzes the performance of the built detector using seven videos
and introduces a new metric to assess the performance of the
detector as automatic detection system. Finally the
 conclusions are summarized in Section~6.

\section{Related works}
The problem of   handgun detection in videos using deep learning is
related in part to two broad research areas. The first addresses gun
detection using classical methods and the second focuses on
improving the performance of  object detection using deep CNNs.

\subsection{Gun detection}

The  first and traditional sub-area in gun detection focuses on
detecting concealed handguns in X-ray or millimetric wave images.
The most representative application in this context is luggage
control in airports. The existent methods achieve high accuracies by
using different combinations of feature extractors and  detectors,
either using simple density descriptors~\cite{PREMIO3}, border
detection and pattern matching~\cite{PREMIO4} or using more complex
methods such as cascade classifiers with boosting~\cite{PREMIO5}.
The effectiveness of these methods made them essential in some
specific places. However, they have several limitations. As these
systems are based on metal detection, they cannot detect non
metallic guns.   They are expensive to be used in many places as
they require to be combined with X-ray scanners and Conveyor belts.
They are not precise because they react to all metallic objects.


The second sub-area addresses gun detection in RGB images using
classical methods. The few existent papers essentially apply methods
such as SIFT (Scale-Invariant Feature Transform) y RIFT
(Rotation-Invariant Feature Transform), combined with  Harris
interest point detector or FREAK (Fast Retina
Keypoint)~\cite{PREMIO6,PREMIO7,PREMIO8}. For example, the authors
in~\cite{PREMIO6,PREMIO7} developed an accurate software for pistol
detection in RGB images. However, their method is unable to detect
multiple pistols in the same scene. The used technique consists of
first, eliminating non related objects to a pistol from the
segmented image using K-mean clustering algorithm then, applying
SURF (Speeded Up Robust Features) method for detecting points of
interest. Similarly,  the authors in~\cite{PREMIO8} demonstrated
that BoWSS (Bag of Words Surveillance System) algorithm has a high
potential to detect guns. They first extract features using SIFT,
cluster the obtained functions using K-Means clustering and use  SVM
(Support Vector Machine) for the training. The authors
in~\cite{PREMIO12} addresses riffle detection in RGB images using
SVM (Support Vector Machine).

All the above cited systems are slow, can not be used for constant
monitoring, require the supervision of an operator and can not be
used in open areas.


\subsection{Detection models}

Object detection consists of recognizing the object and  finding its
location in the input image. The existing methods address the
detection problem by reformulating it into a classification problem,
they first train the classifier then during the detection process
they run it on a number of areas of the input image using either the
sliding window approach or region proposals approach.

\begin{itemize}
\item {\bf Sliding window} approach: It is an exhaustive method that considers a large number of candidate windows,  in the order of $10^4$, from the input image. It scans the input image, at all locations and multiple scales, with a window and runs the classifier at each one of the
windows. The most relevant works in this context improve the
performance of the detection by building more sophisticated
classifiers. The Histogram of Oriented Gradients (HOG) based
model~\cite{dalal2005histograms} uses HOG descriptor for feature
extraction to predict the object class in each window. The
Deformable Parts Models (DPM)~\cite{felzenszwalb2010object}, which
is an extension of HOG based model, uses (1) HOG descriptor to
calculate low-level features, (2) a matching algorithm for
deformable part-based models that uses the pictorial
structures~\cite{felzenszwalb2005pictorial} and (3) a discriminative
learning with latent variables (latent SVM). This model provides
very good accuracies for pedestrian detection with a speed of around
0.07fps and 14s/image.

The obtained accuracies using good classifiers under the sliding
window approach are  satisfactory but the detection process can be
too slow to be used in real time.

\item {\bf Region proposals}  approach: Instead of considering all the possible windows of the input image as candidates, this approach
selects actual candidate regions using detection proposal
methods~\cite{hosang2016makes}.  The first detection model that
introduced CNNs under this approach was Region-based CNNs
(R-CNN)~\cite{girshick2014rich}. It  generates around 2000 potential
bounding boxes using the selective search
method~\cite{uijlings2013selective}, warps the obtained regions into
images of the same size then, feeds them to a powerful CNN-based
classifier to extract their features, scores the boxes using SVM,
adjusts the bounding boxes using a linear model, and
 eliminates duplicate detections via a non-max suppression. R-CNN provides good performance on the well know PASCAL-VOC with
 a speed of  40s/image. Fast R-CNN~\cite{girshick2015fast} and
subsequently Faster R-CNN~\cite{ren2015faster} further improve
computation, data access and disk use of R-CNN. Fast R-CNN has a
speed of 0.5 f/s and 2s/image and Faster R-CNN  around 7f/s and 140
ms/image.
\end{itemize}

This work addresses a new solution to the problem of real-time
pistol detection alarm system using deep learning CNN-based
detector. We  develop, evaluate and compare a CNN based classifier
on different new datasets within the sliding window and region
proposals detection based methods.

\section{Deep learning model}
\label{DLM}

VGGNet was the first runner-up in ILSVRC
2014~\cite{simonyan2014very}. It was used to show that the depth of
the network is critical to the performance. The largest VGGNet
architecture, VGG-16, involves 144 million parameters from 16
convolutional layers with very small receptive fields $3\times 3$,
five max-pooling layers of size $2\times 2$ , three fully-connected
layers, and a linear layer with Softmax activation in the output.
This model also uses dropout regularization in the fully-connected
layer and applies ReLU activation to all the convolutional layers.
This CNN has a greater number of parameters compared to AlexNet and
GoogLeNet, which makes it more computationally expensive.

Deep CNNs, such as VGG-16, are generally trained based on the
prediction loss minimization. Let $x$ and $y$ be the input images
and corresponding output class labels, the objective of the training
is to iteratively minimize the average loss defined as

\begin{equation}
J ( w ) =\frac{1}{N}  \sum_{i=1}^N L ( f ( w ; x_i ), y_i ) +
\lambda R ( w )
\end{equation}

\noindent where $N$ is the number of data instances (mini-batch) in
every iteration, $L$ is the loss function, $f$ is the predicted
output of the network depending on the current weights $w$, and R is
the  weight decay with the Lagrange multiplier $\lambda$. We use the
Stochastic Gradient Descent (SGD), which is commonly used in deep
CNNs to update the weights.

\begin{equation}
w_{t +1} = \mu w_t - \alpha  \Delta J ( w_t )
\end{equation}

\noindent where $\mu$ is the momentum weight for the current weights
$w_t$ and $\alpha$ is the learning rate.

The network weights are randomly initialized if the network is
trained from scratch and are initially set to a pre-trained network
weights if fine-tuning the deep model. In this work we have used
fine-tuning VGG-16 and initialized it with the weights of the same
architecture VGG-16 pre-trained on Imagenet database. The
pre-trained VGG-16 model is available through the deep learning
software used in this work, theano with Keras
front-end~\cite{al2016theano,chollet2015keras} and
Caffe~\cite{jia2014caffe}.

\section{Database construction: Towards an equilibrium between false positives and false negatives} 
Automatic pistol detection in videos not only requires minimizing
the number of false positives but also reaching a near real time
detection.  We analyze the performance of the classifier in
combination with two detection methods, the sliding window
(Section~\ref{SWA}) and the region proposals (Section~\ref{RPA}).

Due to the differences between these two approaches, different
optimization model based on databases with different
characteristics, size and classes, are applied in each case. In the
sliding window approach, we address reducing the number of false
positives by increasing the number of classes and thus building four
databases, Database-1, -2, -3 and -4. The characteristics of all the
databases built in this work are summarized in
Table~\ref{tab_db}. 
In the region proposals approach, the detector is directly trained
on region proposals of a new database, Database-5, with richer
contexts.

{\scriptsize
\begin{table}[h]
\centering \caption{Characteristics of the new training- and
test-sets. The training sets are labeled as Database-1, 2, 3 , 4 and
5.} \label{tab_db}
\begin{tabular}{|l||l|l|l|l|}
 \hline
 Database-& \# classes &  total img  & \#   img of pistols & \# rest of img   \\
 \hline
 1 & 2 & 9100& 3990 (guns)& 5110  \\
2 & 2 &  1857  & 751  & 1056 \\
3 &28 & 5470 & 751 & 4716 \\
4 & 102 &9261 & 200  & 9061 \\
5 & 2& 3000 & 3000   & - \\\hline
Test set & 2 &608 & 304 & 304 \\
 \hline
\end{tabular}
\end{table}
}

To evaluate and compare the performance of the classification model
trained  on the five proposed databases, we have built a testset of
608 images distributed into 304 images that contain pistols and 304
images that do not contain pistols. We have used three metrics, {\it
precision}, {\it recall} and {\it F1 measure}, which evaluates the
balance between the {\it precision} and {\it recall}. Where
$$precision=\frac{True~Positives}{True~Positives+False~Positives},$$
$$recall=\frac{True~Positives}{True~Positives+False~Negatives}$$ and
$$F1~meaure=2\times\frac{precision \times recall}{precision
+recall}$$

For implementing the proposed models we used theano with Keras as
front-end~\cite{al2016theano,chollet2015keras} in the sliding window
based approach shown in Section~\ref{SWA} and
Caffe~\cite{jia2014caffe} in the region proposal approach shown in
Section~\ref{RPA}.

\subsection{Sliding window approach}
\label{SWA} This section aims at finding the best combination of
classifier and training dataset within the sliding window detection
approach. We address the minimization of the number of false
positives by building different training datasets and increasing the
number of classes. 

\subsubsection{Two classes}
\begin{table}[h!]
\centering \caption{The results obtained on the testset by the
classification model. } \label{table2_sw}
\begin{tabular}{|l||l|l|l|l|l|l|l|}
 \hline
Database-  & \#TP     & \#FN   & \#TN &\#FP & Prec. & Rec. & F1
meas.  \\ \hline
1          & 32        & 272     &  109  & 191& 22.70\%  & 14,35 \% &   10.53\% \\
2          & 98        & 206     & 293  & 11 & 89,91\%   & 32,24\% & 47,46\% \\
3          & 85        & 219     & 299  &  5  &  94,44\% & 27,96\%  &43,15\%\\
4          & 97        &  207    & 298  & 6   & 94,17\% &  31,91\% &  {\bf 47,67\%} \\
 \hline
\end{tabular}
\end{table}

To guide our design, and as a first approximation, we have built a
preliminary database, Database-1, and consider a broader class of
type gun. Database-1 contains 9100 images extracted from online gun
catalogues, gun use tutorials and gun advertisement (see details in
Table~\ref{tab_db}). We distributed it into 3990 images of class
gun, which includes different types of guns, pistols, machine guns,
rifle, machine guns, grenade, multiple rocket launcher, tank and
5110 images of not-gun class, which contains different kind of
objects.

We have evaluated the VGG-16 based classification model considering
two classes to indicate either the presence or absence of a pistol
in the scene. Table~\ref{table2_sw} provides the number of true
positives, \#TP, the number of false positives, \#FN, the number of
true negatives, \#TN, the number of false positives, \#FP,
precision, recall and F1 measure obtained by the classification
model when  trained on Database-1, 2, 3 and 4 respectively.

\begin{figure}[t]
        \centering
        \captionsetup[subfloat]{labelformat=empty,justification=centering}
        \subfloat[][]{
\includegraphics[width=20mm ]{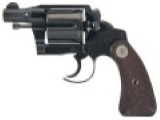}
\includegraphics[width=20mm ]{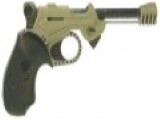}
\includegraphics[width=20mm ]{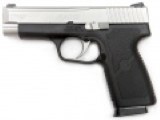}}  \\
       \subfloat[][]{
\includegraphics[width=0.2\textwidth ]{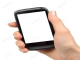}
\includegraphics[width=0.2\textwidth ]{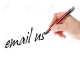}
\includegraphics[width=0.2\textwidth ]{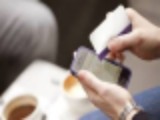} }
        \caption{Examples from Database-2, the top three images  represent the pistol class and the down three images represent the  background class}
        \label{fig:bd2}
\end{figure}

As shown in this Table, the classification model trained on
Database-1 obtains a high number of  false positives, 191,  high
number of false negatives, 272,  and consequently low precision and
recall. This can be explained by the fact that the large variety of
guns makes the learning task very difficult. To make the problem
affordable, we instead focus our task to pistol detection.

In a second step, we have built Database-2 using 1857 images, 751
images of class pistol and 1056 of background. We included to the
background class different images of hands holding different objects
other than pistols, e.g., cell phone and pen as illustrated in
Figure~\ref{fig:bd2}. On the test set, the binomial classification
model trained on Database-2 obtained $11$ false positives, a high
number of false negatives $206$, a precision of $89,91\%$, recall
$32,24\%$ and F1 measure $47,46\%$, which are still below our
expectations. By analyzing the false positives we found that most of
them consider the white background as part of the pistol which is
due to the presence of the white background in most training
examples.

\subsubsection{Multiple classes}

To further decrease the number of false positives we designed two
new databases, Database-3 and Database-4, and considered a higher
number of classes.

Database-3 contains 5470 images set distributed into 28 classes. 751
of  training examples contain pistols. As  shown in
Table~\ref{table2_sw}, the 28-classes based classification model
overcomes the binomial model by reducing the number of false
positives to $5$ and improving the precision to $94,44\%$,
 recall to $27,96\%$  and F1 measure to $43,15\%$. However, the
 number of false negatives is still very high.

We have also explored the performance of the classification model
using a higher number of classes on Database-4. Database-4 is built
using 9261 images distributed into 102 classes, 200 training
examples contain pistols and 9061
 contain diverse objects, e.g., airplanes, ant, beaver, brain, chair, cell phone, animals and so on.
 This
 classifier produces the best results over the previous ones, a lower number of
false negatives $207$, slightly lower number of true negatives $298$
and consequently  better recall $31,91\%$ and F1 measure $47,67\%$.
Next we will evaluate the detection process using the 102-classes
based classifier.

\subsubsection{Detection process and final analysis}
We selected the best classification model to evaluate it under the
sliding window approach. The classifier is applied automatically, in
steps of $60\times 60$-pixels, to windows of $160\times 120$-pixels
of each input image to determine wether it contains a pistol. The
whole detection process takes $1,5$ seconds in a $640\times
360$-pixels input image. Which is not acceptable for an automatic
system such as the one considered in this work.

The detection model that makes use of the sliding window approach
achieves few false positives and high precision, however, it obtains
a low recall 35\% and its execution time is not appropriate for
 online detection. In next section we will explore an other alternative to further improve performance and speed of the
detection process.

\subsection{Region proposals approach}
\label{RPA}
\begin{table}[h!]
\centering \caption{ The results obtained by the classification
model under the region proposals approach on the testset. }
\label{table2_rp}
\begin{tabular}{|l||l|l|l|l|l|l|l|}
 \hline
Database-  & \#TP     & \#FN   & \#TN &\#FP & Prec. & Rec. & F1
meas.  \\ \hline
5          & {\bf 304} & {\bf 0} & {\bf 247}  & 57 & 84,21\% & {\bf 100.00\%} & {\bf 91.43\%} \\
 \hline
\end{tabular}
\end{table}

In this approach, we have used  Faster Region based-CNN (Faster
R-CNN)~\cite{ren2015faster}, which combines the selective search
method with VGG-16 based classifier.

\begin{figure}
\centering {\includegraphics[width=30mm ]{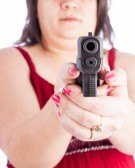}}
{\includegraphics[width=50mm ]{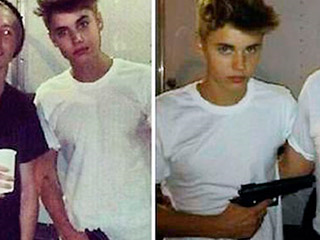}}
{\includegraphics[width=36mm ]{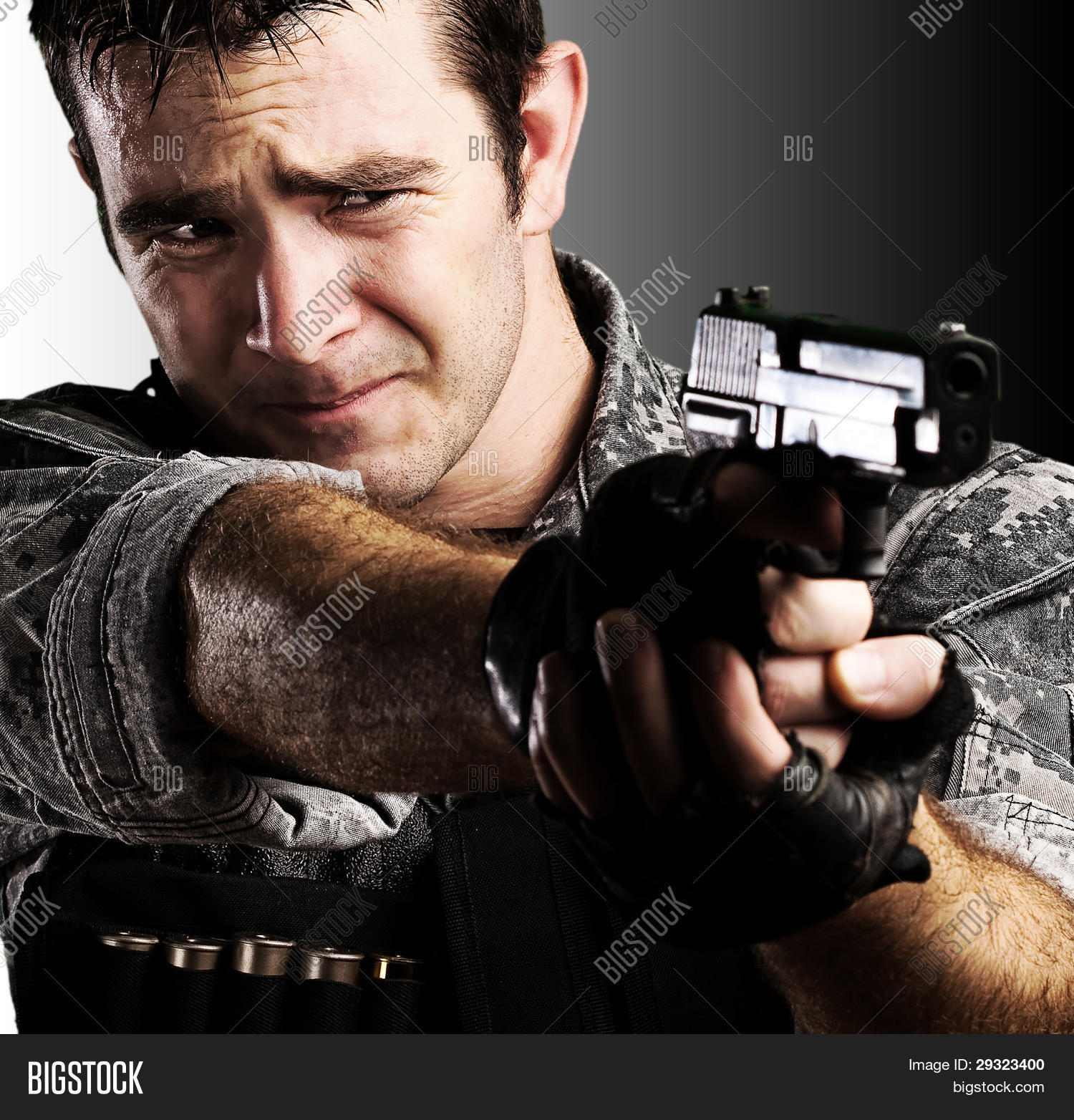}} \caption{Three examples
from Database-5. This dataset includes, in addition to the images,
the xml file of each image with the localization information of the
pistols. } \label{fig:Datab5}
\end{figure}

The design of a new training dataset for this approach is also
manual and cannot re-use the databases from the previous approach.
We have built Database-5 using 3000 images that contains pistols in
different contexts and scenarios, downloaded from diverse web-sites.
Figure~\ref{fig:Datab5} provides three examples of Database-5. We
considered a two class model and labeled the pistols by providing
its localization, i.e., bounding box, in each individual training
image. The rest of objects in the image are considered background.

In general,  as it can be seen from Table~\ref{table2_rp}, Faster
R-CNN  trained on Database-5 obtains the highest performance over
all the previously analyzed models. It provides the highest true
positives number and the highest true negative numbers and
consequently the highest recall $100\%$ and  F1 score $91.43 \%$.
However it produces more false positives, $57$, and consequently
lower precision, $84,21\%$.  In Section~5, we address this issue in
the context of automatic alarm systems by activating the alarm only
when at least five successive false positives happen in five
successive frames.
Next we analyze the speed of this detection model.

\subsubsection{Detection process and final analysis}

We have evaluated the detection process using Faster-RCNN, which is
based on the region proposals approach. We considered two classes
and trained the classification model on Database~5. The whole
detection process in a $1000\times 1000$--pixels image takes $0,19$
seconds approximately producing a rate of $5.3$ frames/s. Which
allows the pistol detection to be performed in videos in near real
time.

The detection model based on region proposals approach achieves the
maximum recall 100\% over the pistol class, zero false negatives,
good precision 85.21\%, a reasonable false positives number and can
be used for real time detection. This makes it a good candidate for
detecting pistols in a sequence of frames as shown in Section 5.

\section{Analysis of the detection in videos}
In this  section we explore the strengths and weaknesses of our
model on seven low quality youtube videos. In particular, we first
assess the quality of the detection and localization
(Section~\ref{Section5-1}) then analyze the suitability of our model
as  pistol detection alarm system using a new metric
(Section~\ref{Section5-2}).

\subsection{Analysis of detection and localization}
\label{Section5-1} This section analyzes the performance of our best
detection model on seven public videos, with low qualities,
downloaded from youtube. Six of them are pieces of well known films
from the 90s, James Bon: The World is Not Enough (video 1), 3 pieces
from Pulp Fiction (video 2, 3 and 4), Impossible Mission: Rogue
Nation (video 5) and Mister Bin (video 6). The seventh video is a
long pistol threatening video (video 7). The videos with the
detections can be found in  a public repository in
github~\footnote{https://github.com/SihamTabik/Pistol-Detection-in-Videos.git}.

\begin{table}
\centering \caption{The total number of True Positives \#TP, total
number of Ground Truth true Positives \#GT\_P, total number of False
Positives \#FP in the considered seven videos, labeled as video 1 to
7 } \label{test_videos}
\begin{tabular}{|l||l|l|l|l|l|l|l|}
\hline video\# & \#frames & \#TP & \#GT\_P &  \#FP &Precision &
Recall & F1 measure \\
\hline 1 & 393 &  60& 162 & 8 &88.24\% & 37.04\% & 52.17\% \\
2 & 627 &  467& 778& 11 & 98.70\% &  60.03\%& 74.36\% \\ 
3 &  441 &  25& 58&  15& 62.50\% &   43.10\%& 51.02\%  \\
4 & 591 & 6 &54 & 0 &100.00\%&      11.11\% & 20.22\% \\
5 & 627 &  24 &105 & 21  &53.33\% &  22.86\%& 32.00\% \\
6 & 212&  141 &290 &30  &82.46\% & 48.62\% & 61.17\% \\ 
7 & 501 &  166 & 476 & 6  &96.51\%&  34.87\% & 51.23\%  \\ 
 \hline
\end{tabular}
\end{table}

\begin{figure}[h]
\centering
{\includegraphics[width=11cm]{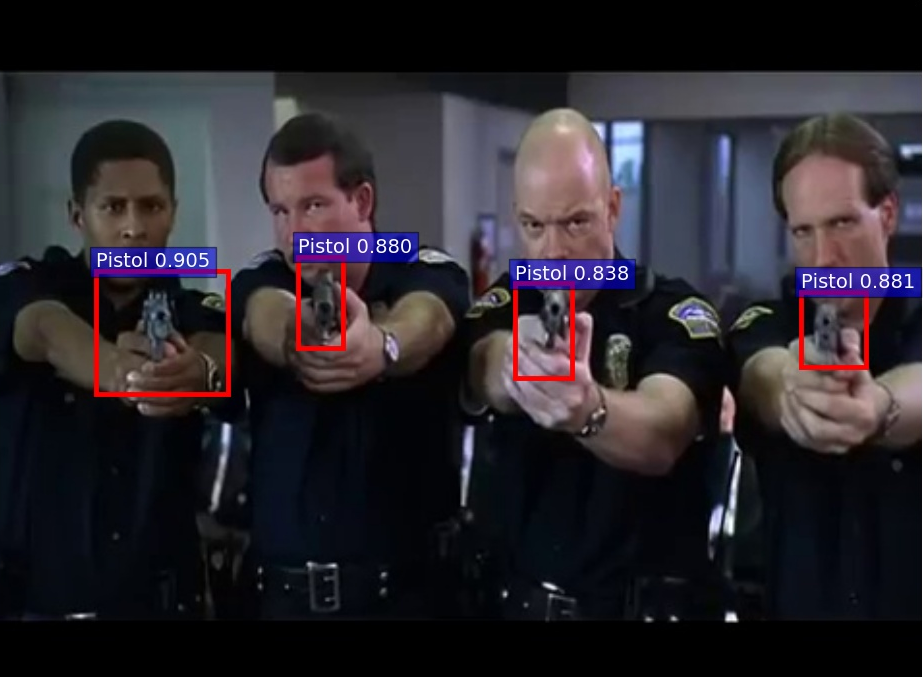}}
\caption{ An example of an accurate detection of four pistols.}
\label{fig:pulp2}
\end{figure}

For the experiments, we analyzed the results of the detection in the
videos frame by frame and  consider a detection as true positive if
the overlapping between the handled pistol and the predicted
bounding box is more than 50\%. Recall that the way pistols are
handled is also a key to the detection. We consider a pistol as
ground truth when it is recognizable by the human eye.
Table~\ref{test_videos} provides the total number of True Positives
\#TP, the total number of False Positives \#FP and total number of
Ground Truth Positives \#GT\_P, in each one of the seven videos. We
consider a threshold $\in [0.7~0.9].$

\begin{figure}[h]
\centering {\includegraphics[width=12cm ]{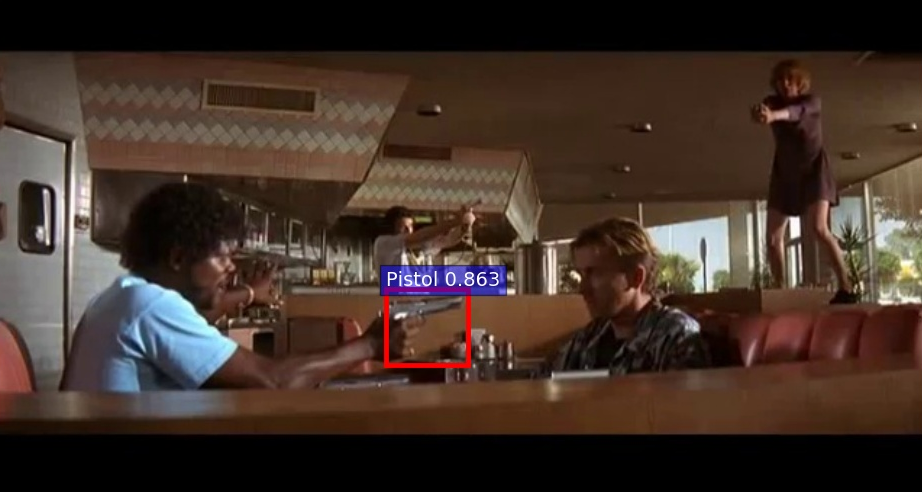}}
\caption{An illustrative example of false negatives, i.e., the two
pistols in background. } \label{fig:pulp1}
\end{figure}
In general, although the scenes are dynamic in most videos, the
detector achieves good balance between precision and recall,
especially in videos 2 and 6. See example of an accurate detection
in Figure~\ref{fig:pulp2}. In particular, the detector provides very
high precisions in videos 1, 2, 3, 4 ,6 and 7, and the obtained
number of false positives is very low in all the videos. Which is
essential to avoid activating negative alarms. The obtained false
positive detections can be addressed in a realistic system by
activating the alarm only when false positives are detected in a
number of consecutive frames.

\begin{figure}[h]
\centering
{\includegraphics[width=5cm ]{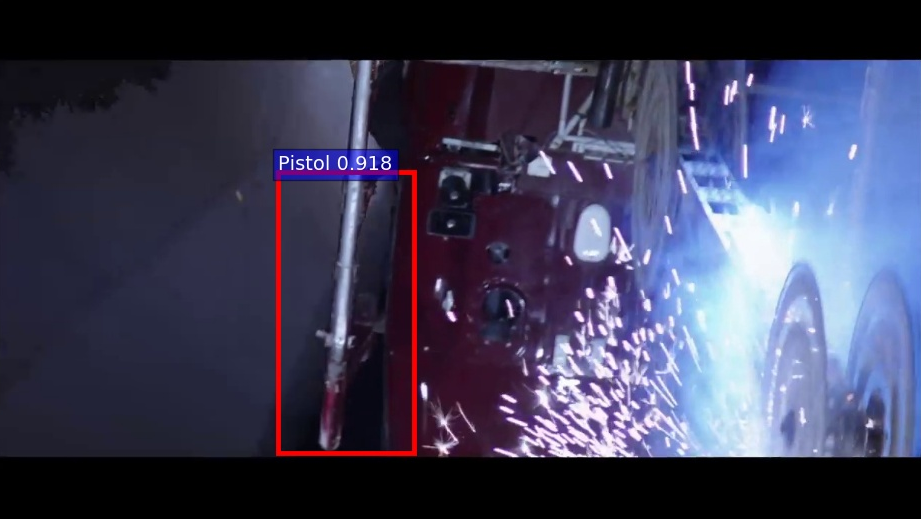}}
{\includegraphics[width=5cm ]{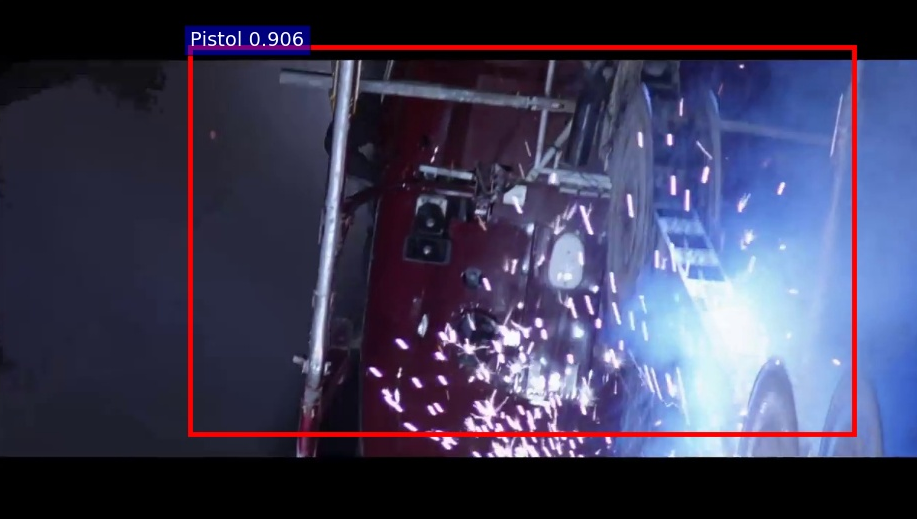}}
\caption{Examples of false positives from Video 1. }
\label{fig:gadget}
\end{figure}
The low recall can be explained by the false negatives detected in
the frames with very low contrast and luminosity  as shown in
Figure~\ref{fig:gadget}. The false negatives depends on the quality
of the  frame and whether the pistol is clearly visible. In
particular, they  occur when the pistol is moved very fast or when
it is placed in the background as in the frame shown in
 Figure~\ref{fig:pulp1}.

 In conclusion, the obtained results can be
 considered as acceptable to detect sequences of clearly visible
 pistols as it will be analyzed in next section.

\subsection{Analyzing the  model as alarm detection system}
\label{Section5-2} In an automatic pistol detection system the alarm
must be activated when the system is completely confident about the
presence of pistols in the scene. To assess the performance of our
detection model as an alarm system, we define a new metric, {\it
Alarm Activation Time per Interval} (AATpI). AATpI quantifies how
fast is the system in detecting  pistols in a given scene. AATpI is
the time the automatic detection alarm system takes to activate an
alarm when it detects at least $k$ successive frames of true
positives. For the analysis presented in this section we used $k=5$.

For the experiments, we selected 30 scenes from the previously used
videos with the next requirements. Each scene is made up of at least
5 frames, filmed in a fixed scenario, i.e., in the same place, and
the pistols are clearly visible to a human viewer. These scenes can
be found in a public repository in
github~\footnote{https://github.com/SihamTabik/Pistol-Detection-in-Videos.git}.

The model successfully detects the pistol in 27 scenes with an
average time interval AATpI=0.2 seconds, which is good enough for an
alarm system. The detector fails to detect pistols only in three
scenes. This is due to the same reasons highlighted previously,
which are the low contrast and luminosity of the frames, the pistol
is moved very fast or when the pistol is not in the foreground.

In summary, although we have used low quality videos for the
evaluation, the proposed model has shown good performance and
demonstrated to be appropriate for  automatic pistol detection alarm
systems.

\section{Conclusions and future work}

This work presented a novel automatic pistol detection system in
videos appropriate for both, surveillance and control purposes. We
reformulate this detection problem into the problem of minimizing
false positives and solve it by building the key training data-set
guided by the results of a VGG-16 based classifier, then assessing
the best classification model under two approaches, the sliding
window approach and region proposal approach. The most promising
results have been obtained with Faster R-CNN based model, trained on
our new database,  providing zero false positives, 100\% recall, a
high number of true negatives and good precision 84,21\%. The best
detector has shown a high potential even in low quality youtube
videos and provides very satisfactory results as automatic alarm
system. Among 30 scenes, it successfully activates the alarm after
five successive true positives within an interval of time smaller
than 0.2 seconds, in 27 scenes.

As present and future work, we are evaluating reducing the number of
false positives,  of Faster R-CNN based detector, by preprocessing
the videos, i.e., increasing their contrast and luminosity, and also
by enriching the training set with pistols in motion. We will also
evaluate different CNNs-based classifier such as, GoogLenet and
consider a higher number of classes.


\section*{Acknowledgments}
This work was partially supported by the Spanish Ministry of Science
and Technology under the project TIN2014-57251-P. Siham Tabik was
supported by the Ramon y Cajal Programme (RYC-2015-18136).






%
\bibliographystyle{plain}
\bibliography{biblio}

%

%
%
%

\end{document}